\documentclass[a4paper,twoside]{article}

\usepackage{graphicx}
\usepackage{epsfig}
\usepackage{subcaption}
\usepackage{calc}
\usepackage{amssymb}
\usepackage{amstext}
\usepackage{amsmath}
\usepackage{amsthm}
\usepackage{multicol}
\usepackage{pslatex}
\usepackage{apalike}
\usepackage{algorithm2e}
\usepackage{booktabs}
\usepackage{multirow}
\usepackage{url}
\usepackage[bottom]{footmisc}
\usepackage{SCITEPRESS}     

\begin{document}

\title{Federated Transformer-GNN for Privacy-Preserving Brain Tumor Localization with Modality-Level Explainability}

\author{\authorname{Andrea Protani\sup{1}\sup{2}\orcidAuthor{0009-0000-7278-8659}, Riccardo Taiello\sup{1}\orcidAuthor{0000-0002-9890-9639} Marc Molina Van Den Bosch\sup{1}\sup{3}\orcidAuthor{0000-0003-1890-6188}, Luigi Serio\sup{1}\orcidAuthor{0000-0002-9346-2663}}
\affiliation{\sup{1}European Organization for Nuclear Research, Geneva, Switzerland}
\affiliation{\sup{2}Dept. of Neuroscience, École Polytechnique Fédérale de Lausanne, Lausanne, Switzerland}
\affiliation{\sup{3}BCN Medtech, Dept. of Engineering, Universitat Pompeu Fabra, Barcelona, Spain}
\email{\{andrea.protani, riccardo.taiello, marc.molina.van.den.bosch, Luigi.Serio\}@cern.ch}}


\abstract{Deep learning models for brain tumor analysis require large and diverse datasets that are often siloed across healthcare institutions due to privacy regulations. We present a federated learning framework for brain tumor localization that enables multi-institutional collaboration without sharing sensitive patient data. Our method extends a hybrid Transformer-Graph Neural Network architecture derived from prior decoder-free supervoxel GNNs and is deployed within CAFEIN\textsuperscript{\textregistered}, CERN's federated learning platform designed for healthcare environments. We provide an explainability analysis through Transformer attention mechanisms that reveals which MRI modalities drive the model predictions. Experiments on the BraTS dataset demonstrate a key finding: while isolated training on individual client data triggers early stopping well before reaching full training capacity, federated learning enables continued model improvement by leveraging distributed data, ultimately matching centralized performance. This result provides strong justification for federated learning when dealing with complex tasks and high-dimensional input data, as aggregating knowledge from multiple institutions significantly benefits the learning process. Our explainability analysis, validated through rigorous statistical testing on the full test set (paired t-tests with Bonferroni correction), reveals that deeper network layers significantly increase attention to T2 and FLAIR modalities ($p<0.001$, Cohen's $d$=1.50), aligning with clinical practice.}

\onecolumn \maketitle \normalsize \setcounter{footnote}{0} \vfill

\section{\uppercase{Introduction}}
\label{sec:introduction}

Brain tumors represent a significant diagnostic challenge requiring accurate localization for treatment planning~\cite{menze2015brats}. Multimodal magnetic resonance imaging (MRI) provides complementary information through T1-weighted, T1-weighted contrast-enhanced (T1ce), T2-weighted, and T2-FLAIR sequences, each capturing distinct tissue characteristics essential for tumor delineation~\cite{bakas2018brats}.

While deep learning has achieved remarkable success in brain tumor segmentation~\cite{isensee2021nnu}, developing robust models requires large, diverse training datasets. However, medical imaging data is inherently distributed across healthcare institutions, and sharing patient data raises significant privacy, legal, and ethical concerns~\cite{rieke2020federated}. This data fragmentation limits the development of generalizable AI models and hinders multi-institutional research collaboration.

Federated learning (FL) addresses this challenge by enabling collaborative model training without centralizing sensitive data~\cite{mcmahan2017fedavg}. Rather than sharing patient records, institutions exchange only model parameters, preserving patient privacy while benefiting from collective learning. This paradigm has shown promise in medical imaging applications~\cite{sheller2020federated}, though challenges remain regarding model interpretability and the impact of data heterogeneity across institutions.

Equally critical for clinical adoption is model explainability. Clinicians require transparent AI systems that provide insights into decision-making processes. Attention-based mechanisms in Transformers offer a natural way to understand which input features contribute most to predictions, enabling interpretation of how different MRI modalities influence tumor detection.

Building upon previous work on decoder-free supervoxel GNNs for brain tumor localization~\cite{protani2025svgformer}, this paper makes two primary contributions:

\begin{enumerate}
    \item \textbf{Federated Learning Framework:} We deploy our Transformer-GNN architecture using CAFEIN\textsuperscript{\textregistered}~\cite{cafein,santos2024federated}, CERN's federated learning platform designed for healthcare applications, demonstrating that federated training significantly outperforms isolated learning on individual client data. While isolated training triggers early stopping due to limited data capacity, federated learning enables continued model improvement, ultimately matching centralized performance. This provides compelling evidence for federated approaches when handling complex medical imaging tasks.

    \item \textbf{Modality-Level Explainability:} We analyze Transformer attention patterns with rigorous statistical validation (ANOVA, paired t-tests with Bonferroni correction, full test set) to reveal which MRI modalities the model prioritizes for tumor detection. We demonstrate statistically significant layer-dependent attention shifts toward T2 and FLAIR modalities ($p<0.001$, Cohen's $d$=1.50), providing clinically meaningful insights that align with radiological practice.
\end{enumerate}

\section{\uppercase{Related Work}}
\label{sec:related}

\subsection{Deep Learning for Brain Tumor Segmentation}
\label{sec:dl_brats}

The Brain Tumor Segmentation (BraTS) challenge has driven significant advances in automated tumor delineation~\cite{menze2015brats,bakas2018brats}. Encoder-decoder architectures dominate current approaches, with nnU-Net~\cite{isensee2021nnu} establishing strong baselines through automated architecture search and preprocessing optimization. Myronenko~\cite{myronenko2019brats} introduced autoencoder regularization to improve generalization, winning the BraTS 2018 challenge.

The emergence of Vision Transformers (ViT)~\cite{dosovitskiy2021vit} has catalyzed new approaches in medical imaging~\cite{shamshad2023transformers}. TransUNet~\cite{chen2021transunet} pioneered the combination of CNN encoders with Transformer layers for medical image segmentation. Swin-UNETR~\cite{hatamizadeh2022swinunetr} adapted hierarchical Swin Transformers specifically for brain tumor segmentation, achieving state-of-the-art results on BraTS benchmarks. TransBTS~\cite{wang2021transbts} demonstrated effective multimodal fusion through Transformer attention mechanisms.

\subsection{Graph Neural Networks for Medical Imaging}
\label{sec:gnn_medical}

Graph-based representations offer advantages for modeling irregular anatomical structures~\cite{parisot2018disease}. Supervoxel methods reduce computational complexity while preserving boundary information: SLIC~\cite{achanta2012slic} provides efficient clustering, while content-aware approaches better capture tissue boundaries~\cite{soltaninejad2018supervoxel}. Graph Attention Networks, particularly GATv2~\cite{brody2022gatv2}, enable dynamic attention over graph neighborhoods, with Laplacian positional encodings capturing topological structure~\cite{dwivedi2023benchmarking}. Recent approaches have combined patch-level Transformers with supervoxel-level GNNs, demonstrating that decoder-free architectures can achieve competitive localization performance.

\subsection{Federated Learning in Healthcare}
\label{sec:fl_healthcare}

Federated learning enables privacy-preserving collaboration across institutions~\cite{rieke2020federated,kairouz2021advances}. McMahan et al.~\cite{mcmahan2017fedavg} introduced FedAvg for communication-efficient distributed training, where clients perform local optimization and periodically synchronize with a central server. The standard FedAvg protocol aggregates client model updates through weighted averaging:
\begin{equation}
\mathbf{w}^{(t+1)} = \sum_{k=1}^{K} \frac{n_k}{\sum_{j=1}^{K} n_j} \mathbf{w}_k^{(t+1)}
\end{equation}
where $\mathbf{w}_k^{(t+1)}$ represents the updated weights from client $k$ after local training on $n_k$ samples. Theoretical analysis established convergence guarantees under non-IID distributions~\cite{li2020convergence}, while FedProx~\cite{li2020fedprox} addressed heterogeneity through proximal regularization. For brain tumor analysis, Sheller et al.~\cite{sheller2020federated} demonstrated that federated models can match centralized performance while preserving patient privacy, motivating multi-institutional collaboration without data sharing. Recent work has further validated MQTT-based decentralized FL frameworks for brain tumor segmentation in clinical settings~\cite{tedeschini2022decentralized}.

\subsection{Explainability in Medical AI}
\label{sec:xai}

Clinical adoption of AI requires transparent decision-making~\cite{salahuddin2022xai}. For convolutional neural networks, Grad-CAM~\cite{selvaraju2020gradcam} has become the de facto standard for generating visual explanations through gradient-weighted class activation mapping. Recent work has investigated explainability under differential privacy constraints in federated settings, revealing that privacy-preserving noise can significantly degrade explanation fidelity, particularly for heterogeneous clients~\cite{molina2025interplay}.

For Transformer architectures, the self-attention mechanism~\cite{vaswani2017attention} provides inherent interpretability by revealing which input tokens influence predictions. In multimodal medical imaging, analyzing attention weights from the CLS token quantifies each modality's contribution, enabling validation against clinical knowledge about imaging sequence informativeness for specific diagnostic tasks.

\section{\uppercase{Methods}}
\label{sec:methods}

\subsection{Supervoxel Graph Representation}
\label{sec:supervoxel}

\begin{figure*}[t!]
  \centering
  \includegraphics[width=\textwidth]{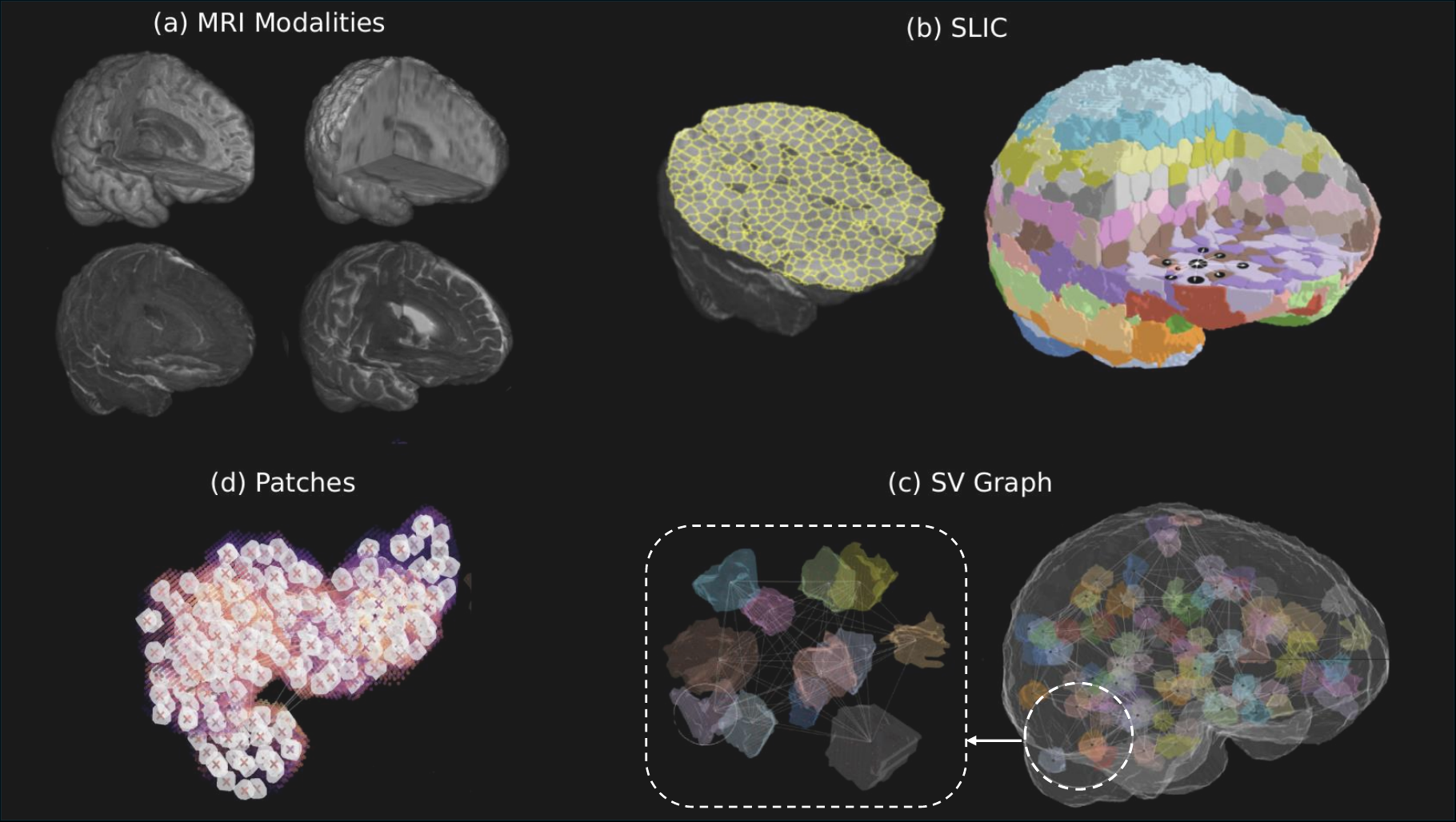}
  \caption{Supervoxel graph preprocessing pipeline: (a) Input MRI modalities; (b) 3D SLIC supervoxel generation applied to T1; (c) Graph construction connecting each supervoxel to its k-nearest neighbors; (d) Patch extraction within each supervoxel for multimodal feature encoding.}
  \label{fig:pipeline}
\end{figure*}

We adopt the supervoxel-based graph representation introduced in prior work~\cite{protani2025svgformer}, illustrated in Figure~\ref{fig:pipeline}. Each BraTS volume comprises four co-registered MRI modalities (T1, T1ce, T2, FLAIR) stored at native resolution, along with voxel-level tumor segmentation masks. Our preprocessing converts raw multi-modal data into compact supervoxel graphs through the following steps:

\textbf{Supervoxel Generation:} Each normalized T1 volume is partitioned into 4000 locally uniform regions using 3D SLIC clustering~\cite{achanta2012slic}. The T1 modality provides stable anatomical contrast, and the resulting supervoxel map is applied uniformly across all co-registered modalities to ensure anatomical consistency.

\textbf{Background Pruning:} Supervoxels with mean T1 intensity below a data-driven threshold are discarded to remove background regions. The threshold is computed by identifying the largest gap in the sorted intensity distribution.

\textbf{Graph Construction:} For each retained supervoxel, we compute its centroid as the mean coordinate of its constituent voxels. Nodes are connected to their $k$=8 nearest neighbors based on Euclidean distance, forming a sparse adjacency structure that captures spatial relationships between anatomically coherent regions.

\textbf{Patch Extraction:} Within each supervoxel, we use k-means++ to select patch centroids. For each centroid, we extract the 45 nearest-neighbor voxel intensities and append 3 spatial coordinates, yielding 48 features per patch. With 90 patches per modality across 4 modalities, each supervoxel is represented by a tensor $\mathbf{X}_i \in \mathbb{R}^{360 \times 48}$.

\textbf{Label Assignment:} Binary classification labels are derived by computing the fraction of voxels within each supervoxel belonging to any tumor class and thresholding at $\tau$=0.20. This threshold was selected based on sensitivity analysis conducted in prior work, which evaluated thresholds from 0.05 to 0.50 and found that $\tau$=0.20 provides the best trade-off between label noise reduction and preservation of small tumor regions.

\textbf{Task Formulation:} Our approach performs tumor \emph{localization} rather than voxel-level segmentation. Each supervoxel is classified as tumor or non-tumor, producing a spatially coherent region that delineates the tumor area. Unlike traditional bounding box localization, our supervoxel-based output provides an anatomically-informed shape that better conforms to irregular tumor boundaries. We report Dice score as an auxiliary metric to quantify spatial overlap between predicted and ground truth tumor regions.

\subsection{Modality-Level Explainability}
\label{sec:explainability}

We extract CLS token attention weights to quantify modality importance. For each node, we compute the attention from the CLS token to patches of each modality, averaged across attention heads within each layer $l$:
\begin{equation}
\text{Attention}_{l,m} = \frac{1}{H} \sum_{h=1}^{H} \frac{1}{|P_m|} \sum_{j \in P_m} \alpha^{(l,h)}_{cls,j}
\end{equation}
where $H$ is the number of attention heads and $P_m$ denotes patches belonging to modality $m$.

\textbf{Statistical Analysis:} To rigorously validate modality attention patterns, we analyze the full test set. For each layer, we perform one-way ANOVA to test whether attention differs across modalities. We then conduct pairwise paired t-tests comparing modality groups (T2+FLAIR vs T1+T1ce), applying Bonferroni correction for multiple comparisons. Effect sizes are reported using Cohen's $d$. A trend test compares the attention difference between Layer 2 and Layer 0 to assess whether deeper layers preferentially attend to clinically relevant modalities.

\section{\uppercase{Experiments}}
\label{sec:experiments}

\subsection{Dataset}
\label{sec:dataset}

We evaluate on the BraTS 2021 dataset~\cite{menze2015brats,bakas2018brats}, which contains 1251 multi-modal MRI volumes with expert-annotated tumor segmentation masks. Each volume includes four co-registered sequences: T1-weighted, T1 contrast-enhanced (T1ce), T2-weighted, and T2-FLAIR. Ground truth masks delineate three tumor sub-regions: enhancing tumor core, non-enhancing core, and peritumoral edema. For our localization task, we binarize these annotations into tumor versus non-tumor labels, merging all sub-regions into a single positive class.

Following our preprocessing pipeline, each volume is converted into a supervoxel graph with approximately 4000 nodes (after background pruning, typically 2500-3500 nodes remain per scan). Each node is represented by 360 patches with 48 features, capturing local tissue characteristics across all modalities. Binary labels mark supervoxels with $>$20\% tumor voxel overlap as positive.

\subsection{Model Architecture}
\label{sec:architecture}

Our architecture comprises three components optimized for federated deployment. Compared to the original architecture~\cite{protani2025svgformer}, we reduce model complexity by 62\% (from 39M to 15M parameters) through reduced embedding dimensions and fewer layers, enabling efficient communication in federated settings while maintaining strong performance.

\textbf{Node Embedder:} Raw patch features are linearly projected into a 192-dimensional space, summed with learnable modality embeddings, and prefixed with a CLS token. A 3-layer Transformer encoder with 6 attention heads processes this sequence. The final node representation concatenates the CLS token output with pooled patch embeddings.

\textbf{Graph Encoder:} A 3-layer GATv2 network~\cite{brody2022gatv2} with 6 attention heads refines node features by modeling inter-supervoxel relationships. Laplacian eigenvector positional encodings (16 dimensions)~\cite{dwivedi2023benchmarking} capture graph topology. Multi-scale fusion aggregates outputs from all layers:
\begin{equation}
\mathbf{f}_i = \text{LayerNorm}(\mathbf{W}_{fuse}[\mathbf{e}^{(1)}_i \| \mathbf{e}^{(2)}_i \| \mathbf{e}^{(3)}_i])
\end{equation}

\textbf{Classifier:} A three-layer MLP with GELU activations and dropout predicts binary tumor probability per node. We employ a compound loss combining Focal Loss~\cite{lin2017focal}, Dice Loss~\cite{milletari2016vnet}, and a recall-focused term to address class imbalance while prioritizing tumor detection sensitivity.

\subsection{Federated Training Configuration}
\label{sec:fed_config}

We simulate 4 institutional clients with varying dataset sizes (18\%, 22\%, 35\%, 25\% of total samples). The unequal partitioning allows us to study the effect of different local training set sizes on federated learning dynamics, reflecting realistic scenarios where institutions contribute different volumes of data.

\textbf{Communication Efficiency:} The 62\% parameter reduction from 39M to 15M parameters directly translates to reduced communication overhead. With 15M parameters stored in 32-bit precision, each model update requires approximately 60 MB per communication round. Using half-precision (BF16) for weight transfer reduces this to 30 MB. Over 600 communication rounds, total data transfer amounts to approximately 18 to 36 GB per client, which is well within the capacity of modern hospital network infrastructure and can be efficiently handled by lightweight messaging protocols. This communication footprint is comparable to transferring a few high-resolution CT scans, making federated training practical for healthcare institutions with standard network connectivity.

\subsection{Experimental Setup}
\label{sec:exp_setup}

We compare three training paradigms to evaluate the impact of data aggregation:

\textbf{Centralized Training:} All 1251 samples are pooled and split 80/20 for training/test. This serves as the upper-bound baseline representing unlimited data access.

\textbf{Federated Training:} Data is distributed across 4 simulated institutional clients with varying sizes: Client 1 (18\%, 225 samples), Client 2 (22\%, 275 samples), Client 3 (35\%, 438 samples), and Client 4 (25\%, 313 samples). Each client maintains a local 80/20 train/test split. The server aggregates model updates using FedAvg after each communication round. We experimented with two local training strategies: fixed step count per client versus full local epoch; the latter yielded better convergence and was adopted for all experiments.

\textbf{Isolated Training:} Each client trains independently using only their local data, with no parameter sharing. This baseline quantifies the benefit of federated collaboration over siloed institutional training.

All paradigms use identical model architectures (15M parameters) and hyperparameters, differing only in data access patterns.

\subsection{Training Details}
\label{sec:training}

Models are implemented in PyTorch 2.2 and PyTorch Geometric 2.5, trained on NVIDIA A100 GPUs (40GB). The architecture comprises 15M parameters distributed across the Node Embedder (2.5M), Graph Encoder (11M), and Classifier (1.5M).

Training employs the AdamW optimizer with learning rate $3 \times 10^{-5}$ and weight decay $10^{-2}$, using cosine annealing with warm restarts. We apply gradient clipping (max norm 1.0), dropout (p=0.2), and mixed-precision training (BF16) for efficiency. Batch size is 2 with gradient accumulation over 8 steps.

Centralized training runs for 200 epochs. Federated training performs 600 communication rounds with 1 local epoch per round. All paradigms employ early stopping with patience of 100 epochs/rounds, monitoring test Dice score.

\section{\uppercase{Results}}
\label{sec:results}

\subsection{Federated vs. Centralized vs. Isolated Learning}

Table~\ref{tab:results} summarizes the test set performance across all training paradigms. To ensure fair comparison, the test set for centralized and federated training is constructed by aggregating the local test sets from all clients. Federated learning achieves performance statistically indistinguishable from centralized training, with overlapping confidence intervals (Dice $0.60\pm0.02$ vs $0.59\pm0.02$). In contrast, isolated training yields substantially lower scores (Dice $0.54\pm0.03$). Among isolated clients, Client 2 (22\% of data) achieves the highest Dice (0.57), followed by Client 3 (35\%, Dice 0.55), Client 1 (18\%, Dice 0.53), and Client 4 (25\%, Dice 0.49). While the differences between individual clients fall within overlapping confidence intervals and should be interpreted cautiously, the overall pattern suggests that dataset size is not the sole determinant of isolated performance. The relatively narrow performance range across clients (Dice 0.49 to 0.57) highlights the fundamental limitation of isolated training regardless of local dataset size.

\begin{table}[ht]
\centering
\caption{Test set performance across training paradigms. Values show mean $\pm$ std over 3 runs with different random seeds. Isolated Avg shows mean $\pm$ std across the 4 clients.}
\label{tab:results}
\footnotesize
\setlength{\tabcolsep}{2pt}
\begin{tabular*}{\columnwidth}{@{\extracolsep{\fill}}lcccc@{}}
\toprule
\textbf{Paradigm} & \textbf{Dice} & \textbf{Prec.} & \textbf{Rec.} & \textbf{F1} \\
\midrule
Centralized & 0.59\tiny{$\pm$.01} & 0.78\tiny{$\pm$.02} & 0.84\tiny{$\pm$.01} & 0.80\tiny{$\pm$.01} \\
Federated & 0.60\tiny{$\pm$.02} & 0.78\tiny{$\pm$.01} & 0.85\tiny{$\pm$.02} & 0.81\tiny{$\pm$.01} \\
\midrule
Isolated Avg & 0.54\tiny{$\pm$.03} & 0.73\tiny{$\pm$.02} & 0.82\tiny{$\pm$.01} & 0.77\tiny{$\pm$.02} \\
\quad Client 1 & 0.53\tiny{$\pm$.02} & 0.71\tiny{$\pm$.02} & 0.82\tiny{$\pm$.01} & 0.75\tiny{$\pm$.01} \\
\quad Client 2 & 0.57\tiny{$\pm$.01} & 0.75\tiny{$\pm$.01} & 0.82\tiny{$\pm$.02} & 0.78\tiny{$\pm$.01} \\
\quad Client 3 & 0.55\tiny{$\pm$.02} & 0.75\tiny{$\pm$.02} & 0.83\tiny{$\pm$.01} & 0.78\tiny{$\pm$.01} \\
\quad Client 4 & 0.49\tiny{$\pm$.03} & 0.72\tiny{$\pm$.02} & 0.80\tiny{$\pm$.02} & 0.75\tiny{$\pm$.02} \\
\bottomrule
\end{tabular*}
\end{table}

\begin{figure*}[t!]
  \centering
  \includegraphics[width=\textwidth]{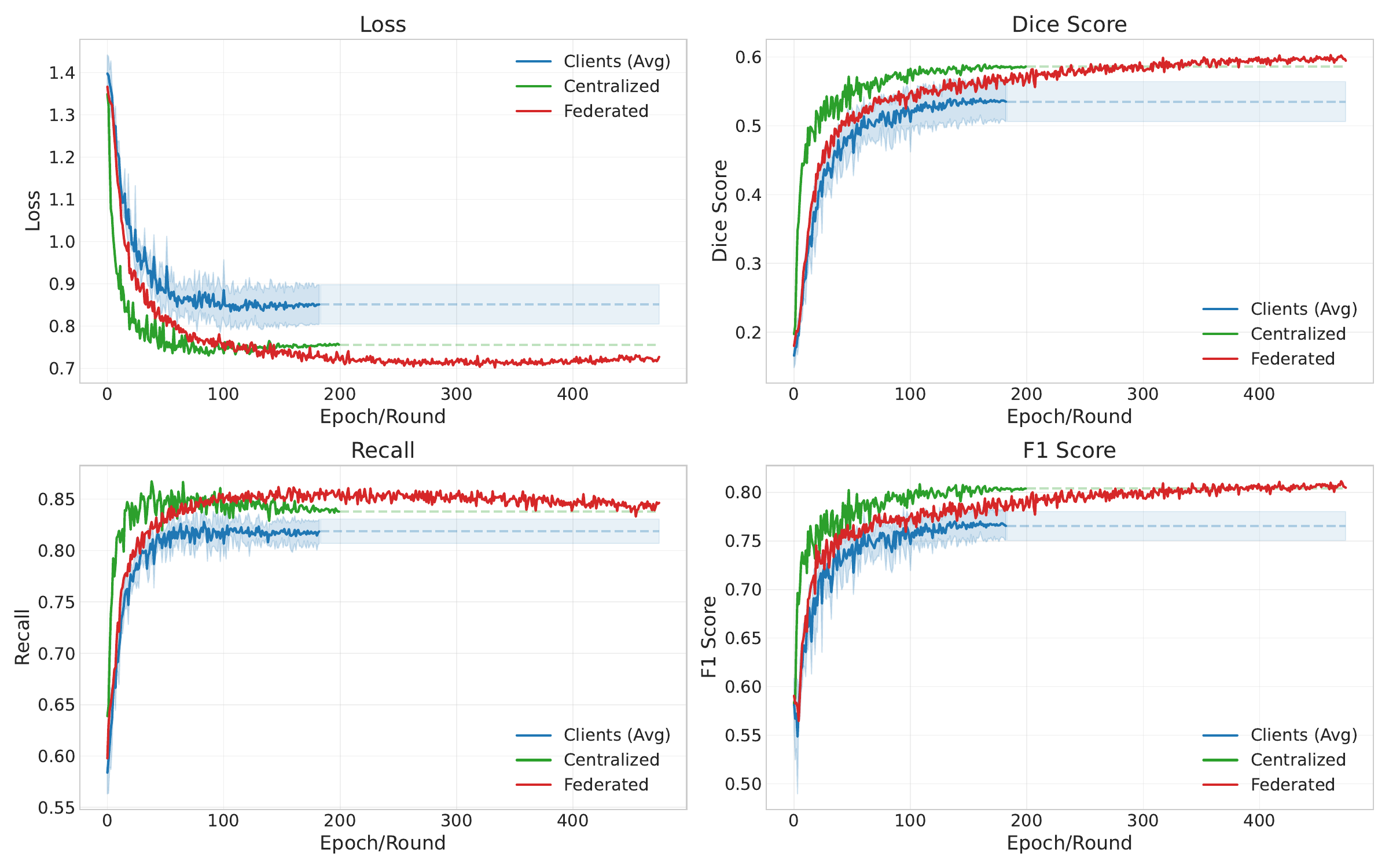}
  \caption{Training dynamics comparison between centralized, federated, and isolated (average across clients) learning paradigms. Evolution of loss, Dice score, recall, and F1 score over training epochs/rounds. Shaded bands represent variability across clients. For visualization purposes, centralized and isolated curves are extended beyond their training duration using the last recorded value (dashed lines with reduced opacity) to facilitate comparison with federated training. Key observation: isolated training plateaus early due to limited local data, while federated learning continues improving and ultimately matches centralized performance.}
  \label{fig:comparison}
\end{figure*}

Figure~\ref{fig:comparison} presents the training dynamics comparing centralized, federated, and isolated learning paradigms. The most significant finding emerges from comparing isolated versus federated training:

\textbf{Isolated Learning Limitations:} When training independently on local client data, models trigger early stopping well before reaching 200 epochs. The limited data available at each institution is insufficient for the model to continue learning, causing test performance to plateau and early stopping to activate. This demonstrates that individual institutional datasets, while valuable, cannot fully support the complexity of the Transformer-GNN architecture.

\textbf{Federated Learning Advantage:} In contrast, federated training enables continued model improvement throughout the training process. By aggregating knowledge from all participating institutions, the model accesses a larger and more representative training set. Although federated convergence is slower than centralized training due to communication overhead, the federated model ultimately matches centralized performance without triggering early stopping prematurely.

\textbf{Justification for Federated Learning:} This result provides strong justification for adopting federated learning when dealing with complex tasks and high-dimensional input data. The aggregation of distributed knowledge proves essential for achieving optimal model capacity. Even with slower convergence, the ability to leverage data from multiple sources without direct sharing enables models to learn representations that would be impossible with isolated institutional training.

\textbf{Recall Preservation:} Federated training maintains high sensitivity throughout training, suggesting that FedAvg aggregation effectively preserves the model's ability to detect tumor regions across diverse institutional datasets. This is critical for clinical applications where false negatives carry significant consequences.

These results demonstrate the practical value of privacy-preserving collaboration for complex medical imaging tasks.

\subsection{Explainability Analysis}

\begin{figure*}[t!]
  \centering
  \includegraphics[width=0.8\textwidth]{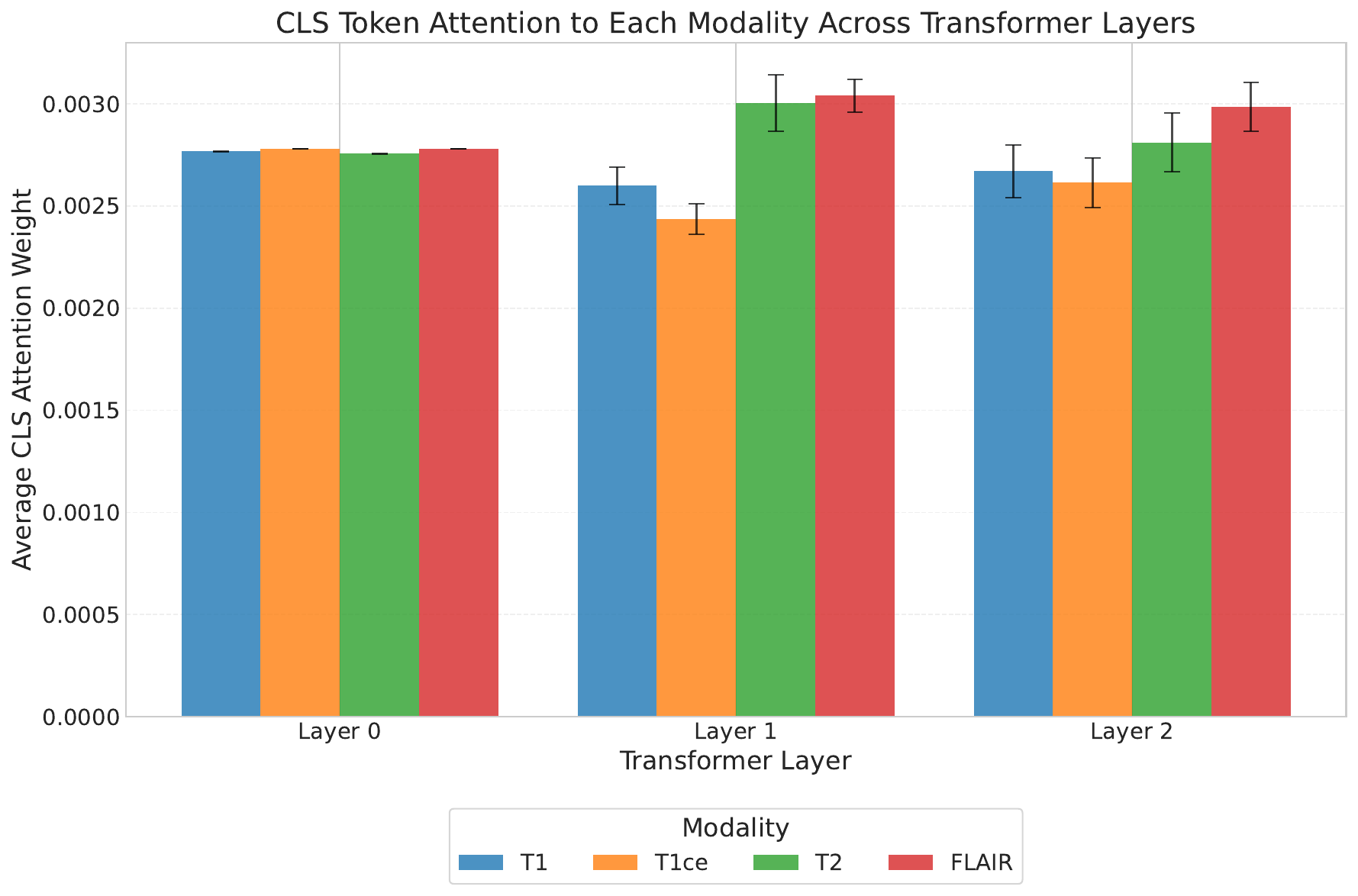}
  \caption{CLS token attention distribution across modalities and Transformer layers (full test set, error bars $\pm$1 SD). Statistical analysis reveals that deeper layers significantly increase attention to T2 and FLAIR modalities compared to early layers (trend test: $p<0.001$, Cohen's $d$=1.50).}
  \label{fig:interpretability}
\end{figure*}

Figure~\ref{fig:interpretability} illustrates the modality attention patterns across Transformer layers. We performed rigorous statistical analysis using one-way ANOVA within each layer, followed by paired t-tests with Bonferroni correction for pairwise comparisons (full test set).

\textbf{Layer-Dependent Modality Emphasis:} One-way ANOVA revealed significant differences in modality attention within all layers ($p<0.001$). Critically, comparing combined T2+FLAIR versus T1+T1ce attention shows a clear layer-dependent pattern: Layer 0 exhibits uniform attention across modalities (difference: $-0.2\%$, $p=1.0$), while deeper layers show significantly elevated attention to T2 and FLAIR sequences. Layer 1 exhibits the strongest preference ($+20.1\%$, $p<0.001$, Cohen's $d$=3.70), with Layer 2 maintaining significant preference ($+9.7\%$, $p<0.001$, Cohen's $d$=1.47). A formal trend test confirms that the Layer 0 to Layer 2 shift is statistically significant ($p<0.001$, Cohen's $d$=1.50). This emergent behavior aligns with clinical knowledge that T2 and FLAIR are most informative for identifying edema and tumor boundaries~\cite{bakas2018brats}.

\textbf{Complementary Information Utilization:} Pairwise comparisons reveal that in deeper layers, FLAIR receives significantly more attention than T1ce ($p<0.001$), while T2 receives significantly more than T1 ($p<0.01$). Interestingly, T2 and FLAIR attention do not significantly differ in Layer 1 ($p=0.30$), suggesting the model treats these modalities as complementary for tumor-related features. This validates that the network captures clinically meaningful multimodal relationships.

\textbf{Consistency Across Samples:} The relatively small standard deviations across the test set indicate robust modality-specific feature extraction. The mean per-patient correlation between layer depth and T2+FLAIR preference is $r$=0.45 ($p<0.001$), confirming consistent attention patterns rather than overfitting to individual cases.

\section{\uppercase{Discussion}}
\label{sec:discussion}

Our results demonstrate that federated learning provides substantial benefits over isolated training for brain tumor localization, particularly when dealing with complex architectures and high-dimensional multimodal data. The key finding that isolated training triggers early stopping while federated training continues improving has important implications for medical AI deployment.

This observation suggests that for sophisticated models like Transformer-GNN architectures, the data available at individual institutions may be insufficient to fully exploit the model's learning capacity. Federated learning addresses this limitation by enabling knowledge aggregation across institutions without compromising patient privacy. The slower convergence of federated training, often cited as a drawback, is far preferable to the premature convergence observed in isolated settings.

The modality-level explainability analysis addresses a critical requirement for clinical AI adoption. Statistical testing with Bonferroni correction (full test set) confirms that deeper layers preferentially attend to T2 and FLAIR with large effect sizes (Cohen's $d$ ranging from 1.47 to 3.70), which are the modalities clinicians prioritize for tumor and edema assessment. The non-uniform attention pattern, with Layer 1 showing peak preference before moderating in Layer 2, suggests the model develops hierarchical feature extraction where middle layers focus most strongly on tumor-discriminative modalities, while the final layer integrates features for classification. This emergent behavior validates the approach and may enhance clinician trust in AI-assisted diagnosis.

\textbf{Clinical Implications:} The alignment between model attention patterns and established radiological practice has significant implications for clinical deployment. Radiologists routinely rely on T2 and FLAIR sequences to assess peritumoral edema extent and distinguish tumor boundaries from surrounding healthy tissue~\cite{bakas2018brats}. Our finding that the model autonomously learns to prioritize these same modalities suggests that the network captures clinically meaningful features rather than exploiting dataset-specific artifacts. This interpretability could facilitate human-AI collaboration, where clinicians can verify that model predictions are based on appropriate imaging features, potentially increasing acceptance of AI-assisted diagnosis in clinical workflows.

\textbf{Limitations:} Our approach focuses on tumor localization rather than pixel-precise segmentation, providing supervoxel-level detection that offers a practical balance between accuracy and computational efficiency. The federated experiments use simulated clients from a single dataset to enable controlled comparison of training paradigms; validation on multi-site clinical deployments with naturally occurring data heterogeneity remains an important next step. Additionally, while federated learning inherently preserves data locality, integration of formal differential privacy mechanisms could further strengthen privacy guarantees for deployment in highly regulated environments.

\textbf{Future Work:} Several directions emerge from this work. We plan to validate our framework on geographically distributed clinical datasets with true inter-institutional data heterogeneity, partnering with hospitals across different countries to assess robustness to domain shift. Integration of differential privacy~\cite{molina2025interplay} within the CAFEIN\textsuperscript{\textregistered} platform would provide formal privacy guarantees, though careful analysis of the privacy-utility trade-off will be essential. We also aim to extend the approach to other brain pathologies (e.g., stroke lesions, multiple sclerosis plaques) and imaging modalities beyond MRI. Finally, prospective clinical validation studies will be necessary to assess real-world utility and clinician acceptance before deployment in diagnostic workflows.

\section{\uppercase{Conclusions}}
\label{sec:conclusion}

We presented a federated learning framework for brain tumor localization that demonstrates the critical importance of collaborative training for complex medical imaging tasks. Our key finding shows that isolated training on individual client data triggers early stopping due to limited data capacity, while federated learning enables continued model improvement, ultimately matching centralized performance. This result provides compelling justification for federated approaches when dealing with sophisticated architectures and high-dimensional input data.

By deploying a hybrid Transformer-GNN architecture in a federated setting, we showed that aggregating knowledge from distributed sources is essential for achieving optimal model capacity while preserving patient privacy. Our modality-level explainability analysis, supported by rigorous statistical testing (full test set, $p<0.001$, Cohen's $d$=1.50), confirms that the model learns clinically meaningful patterns, with deeper layers significantly increasing attention to T2 and FLAIR modalities consistent with radiological practice. These contributions advance the development of trustworthy, privacy-preserving AI for brain tumor analysis.

\section*{\uppercase{Acknowledgements}}

This project was partially funded by the CERN Medical Applications Knowledge Transfer budget.

\bibliographystyle{apalike}
{\small
\bibliography{bibliography}}

\end{document}